\documentclass[letter, 10pt]{IEEEtran}
\usepackage{graphicx}
\usepackage{url}
\usepackage{multicol}
\usepackage{subfig}
\usepackage{booktabs}

\usepackage{amsmath,amssymb}
\usepackage{multirow}
\newcommand{\tabincell}[2]{\begin{tabular}{@{}#1@{}}#2\end{tabular}}
\usepackage[font=small]{caption}
\usepackage{amsmath}
\usepackage{amssymb}
\usepackage{float}
\usepackage{multicol}

\usepackage{lettrine}%  dropped capital letter at beginning of paragraph
\usepackage{algorithm}
\usepackage{algpseudocode}
\usepackage{cite}
\usepackage{filecontents}
\usepackage{lipsum}
\usepackage{color}
\usepackage{esdiff}
\usepackage{epstopdf}
\usepackage{flushend}
\usepackage[utf8]{inputenc}
\usepackage[english]{babel}
\usepackage{amsthm}
\newtheorem{defn}{Definition}[section]

\theoremstyle{remark}

 \newcommand{\blue}{\color{blue}}

\begin{document}

\title{{\small \blue IEEE/MTS Global Oceans 2020, Singapore-US Gulf Coast}\\ \vspace{-3pt}
\Huge {$\epsilon^\star$+: An Online Coverage Path Planning Algorithm for Energy-constrained Autonomous Vehicles}\\
}

%Paper authors and thank you.
\author{ \begin{tabular}{cccccccccc}
{Zongyuan Shen$^\dag$$^\star$} & {James P. Wilson$^\dag$} & {Shalabh Gupta$^\dag$}
\end{tabular}\vspace{0pt}
\thanks {$^\dag$Dept. of Electrical and Computer Engineering, University of Connecticut, Storrs, CT 06269, USA.}
\thanks {$^\star$Corresponding author (email: zongyuan.shen@uconn.edu)}}

\maketitle

\begin{abstract}
This paper presents a novel algorithm, called $\epsilon^{\star}$+, for online coverage path planning of unknown environments using energy-constrained autonomous vehicles. Due to limited battery size, the energy-constrained vehicles have limited duration of operation time. Therefore, while executing a coverage trajectory, the vehicle has to return to the charging station for a recharge before the battery runs out. In this regard, the $\epsilon^\star$+ algorithm  enables the vehicle to retreat back to the charging station based on the remaining energy which is monitored throughout the coverage process. This is followed by an advance trajectory that takes the vehicle to a near by unexplored waypoint to restart the coverage process, instead of taking it back to the previous left over point of the retreat trajectory; thus reducing the overall coverage time. The proposed $\epsilon^\star$+ algorithm is an extension of the $\epsilon^\star$ algorithm, which utilizes an Exploratory Turing Machine (ETM) as a supervisor to navigate the vehicle with back and forth trajectory for complete coverage. The performance of the $\epsilon^\star$+ algorithm is validated on complex scenarios using Player/Stage which is a high-fidelity robotic simulator.
\end{abstract}


\IEEEpeerreviewmaketitle

\thispagestyle{empty}

\vspace{-12pt}
\section{Introduction}
Coverage path planning (CPP) aims to find a trajectory for an autonomous vehicle that enables it to pass over all points in the search area while avoiding obstacles. CPP has a wide range of applications for autonomous underwater vehicles (AUVs), such as seafloor mapping~\cite{shen2016,shen2017,shen2017mastersthesis}, structural inspection~\cite{englot2013}, mine hunting~\cite{mukherjee2011}, oil spill cleaning~\cite{song2013} and other underwater tasks. A variety of methods have been developed to solve the CPP problem~\cite{acar2002,gabriely2003,gonzalez2005}; a review of the existing methods is presented in~\cite{galceran2013}. In general, CPP methods can be categorized into offline or online methods. While offline methods assume the environment to be \textit{a priori} known, online methods~\cite{acar2002,gabriely2003,gonzalez2005,galceran2013,shen2019,song2018} compute the coverage path \textit{in situ} based on sensor information. However, in real-life situations, AUVs are energy-constrained due to the limited size of batteries; thus limiting their duration of operation. Therefore, while executing a coverage trajectory, the AUV has to return to the charging station for a recharge before the battery runs out. While a variety of CPP methods are available in literature, only a limited body of work has focused on energy-constrained CPP problem.

\subsection{Background}
Shnaps et. al.~\cite{shnaps2016} presented an online coverage strategy called battery powered coverage (BPC), using a vehicle with limited energy. This method relies on the concept of equipotential contours, split cells and corridors. Specifically, an equipotential contour is a curve on which each point has the same distance to the charging station. A split cell is the free cell which is adjacent to an obstacle boundary if the equipotential contour splits at this cell into two segments. Corridor is a subregion consisting of a set of equipotential contour segments, which start at a split cell and are bounded by an obstacle. Starting at the charging station, the vehicle follows the nearest contour and advances through farther contours within the same corridor. During navigation, the battery level is monitored and when the battery level is low, the vehicle returns to the charging station along the shortest path, gets full recharge, and continues the coverage of previous corridor; if it has been completely covered, then the vehicle proceeds to cover the alternate corridor which is closest to the vehicle. However, this method has the following disadvantages: i) a contour path is difficult to follow by the vehicle's controller, especially if the localization system has inaccuracies or uncertainties, ii) a contour path is less desirable from the user's perspective, in comparison to the back-and-forth path, iii) after recharge the vehicle returns to the previous corridor instead of starting at some nearby point, and iv) once a corridor is covered, the vehicle goes to the previously stacked split cell, which could be far away from the vehicle's position, and then proceeds to cover the alternate corridor. The last two items can generate redundant paths with longer lengths. %, specially when the split cell is lying far away from the vehicle's position. %ii) the contour path may generate more number of turns than the back-and-forth path, 

Wei et. al.~\cite{wei2018} proposed a similar method for offline coverage of a planar area using from-far-and-near coverage pattern. This method models the environment by a set of equi-distance polylines on which the centroids of the cells have the same distance to the charging station. Note that a polyline is a connected series of line segments, also called a polygonal curve. Starting at the charging station, the vehicle goes to the furthest uncovered cells and follows the corresponding polyline for coverage. Once the current polyline is completely covered, the vehicle moves to the adjacent polyline and continues the coverage process. However, it's performance can degrade if \textit{a priori} knowledge is incomplete or incorrect. 

\vspace{-6pt}
\subsection{Our Approach}

To address the above issues, this paper presents a novel algorithm, called $\epsilon^\star$+,  which is an extension of the $\epsilon^\star$ algorithm~\cite{song2018}, for online CPP of unknown environments using energy-constrained vehicles. During coverage, the vehicle detects unknown obstacles for dynamic map-building, while continuously monitoring the remaining energy so that it has sufficient energy for returning to the charging station when its battery runs low. Upon  battery depletion, the vehicle retreats back to the charging station and after recharging it advances to a nearby cell to restart the coverage of the remaining uncovered area. The performance of the proposed algorithm is validated on a high-fidelity simulator called Player/Stage.

The advantages of the $\epsilon^\star$+ algorithm are as follows: i) produces easy-to-follow and user-desired back-and-forth coverage trajectories in contrast to the circular contours, ii) chooses a nearby unexplored cell as the new start point after each recharging to avoid longer travel to the cell from where it retreated back, iii) guarantees complete coverage with minimum overlaps, iv) works in unknown environments, and v) computationally efficient for real-time implementation.

\vspace{-6pt}
\subsection{Organization}
The rest of the paper is organized as follows. Section~\ref{sec:problem} formulates the energy-constrained CPP problem. Section~\ref{sec:algorithm} describes the $\epsilon^\star$+ algorithm. Section~\ref{sec:results} presents the results and Section~\ref{sec:conclusions} presents the conclusions and future work.

\vspace{-0pt}
\section{Problem Statement}\label{sec:problem}

Let $\mathcal{A} \subset \mathbb{R}^2$ be the desired coverage area. First, we construct a tiling on $\mathcal{A}$ as follows.

\begin{defn}[Tiling]\label{define:tiling}
A set $\mathcal{T}=\{\tau_i \subset \mathbb{R}^2, i=1,\ldots|\mathcal{T}|\}$ is called a tiling of $\mathcal{A}$ if its elements, called tiles (or cells), i) have mutually exclusive interiors, i.e., $\tau^0_{i} \cap \tau^0_{j} =\emptyset, \forall i \neq j$, where $i, j \in \{1,\ldots,|\mathcal{T}|\}$, and superscript $0$ denotes the interior of a cell, and ii) form a cover of $\mathcal{A}$, i.e., $\mathcal{A} \subseteq \bigcup_{i=1}^{|\mathcal{T}|  }\tau_i$
\end{defn}

\begin{figure}[t]
    \centering
    \includegraphics[width=0.65\columnwidth]{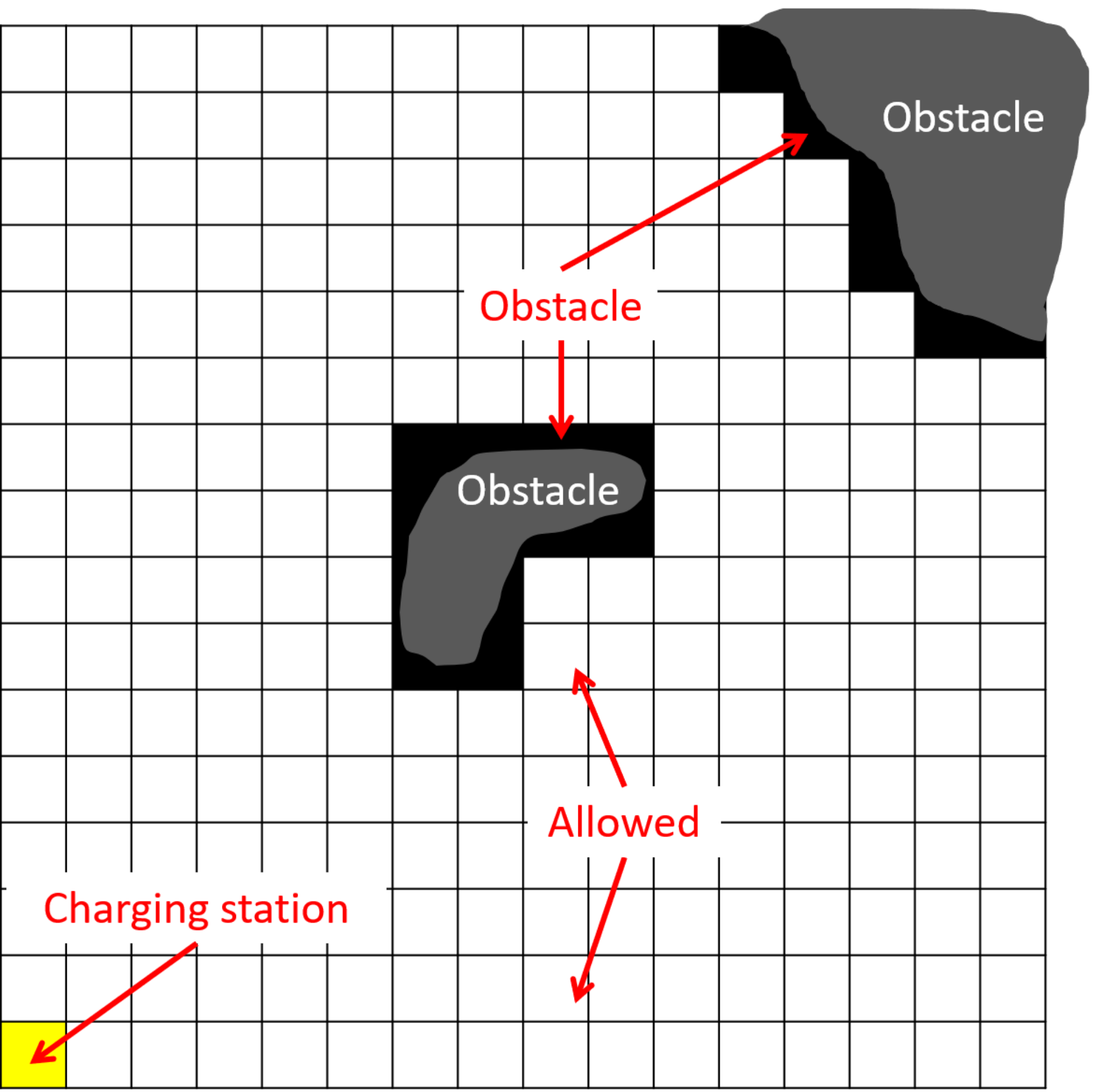}
    \caption{Tiling of the search area}
   \label{fig:tiling} \vspace{-6pt}
\end{figure}

The tiling $\mathcal{T}$ is partitioned into two sets: i) obstacle ($\mathcal{T}^o$) and ii) allowed ($\mathcal{T}^a$), as shown in Fig~\ref{fig:tiling}. The allowed cells are further classified as unexplored and explored (i.e., covered). Now, we discuss the online energy-constrained CPP problem. 

%\begin{defn}[Online Energy-constrained CPP Problem]\label{define:coverage}
Let $\mathcal{A}(\mathcal{T}^a)$ denote the total area of the allowed cells in $\mathcal{T}^a \subseteq \mathcal{T}$. Let $\tau_{C} \in \mathcal{T}$ be the cell at which the charging station is located. Let $E_0\in \mathbb{R}^{+}$ be the total energy that the vehicle has under full charge.

\begin{defn}[Trajectory]\label{define:trajectory}
A trajectory $\pi$ is defined as a sequence of cells visited by the autonomous vehicle consisting of three segments: i) advance, ii) coverage, and iii) retreat. 
\begin{itemize}
\item The advance segment takes the vehicle from the charging station to an unexplored cell, 
\item The coverage segment performs coverage of unexplored cells using the $\epsilon^\star$ algorithm till the vehicle's battery depletes to the extent that allows it to go back to the charging station, and 
\item The retreat segment brings the vehicle back to the charging station along the shortest path.
\end{itemize}
\end{defn}

Let $\mathcal{A}(\pi_n)$ denote the area covered by a trajectory $\pi_n$. Then, the goal of the energy-constrained CPP problem is to find an ordered set of trajectories $\Pi=\{\pi_1,\pi_2\ldots \pi_N\}$ such that:

\begin{itemize}
\item Each trajectory $\pi_n \in \Pi$ starts and ends at $\tau_C$.
\item Each trajectory $\pi_n \in \Pi$ consumes energy $E(\pi_n) \leq E_0$.
\item $\Pi$ forms a cover of $\mathcal{A}(\mathcal{T}^a)$, i.e.,  $\mathcal{A}(\mathcal{T}^a) \subseteq \bigcup_{n=1}^{N}\mathcal{A}(\pi_n)$. 
\end{itemize}

%\end{defn}

\section{$\epsilon^\star$+ algorithm}\label{sec:algorithm}
The $\epsilon^\star$+ algorithm is an extension of the $\epsilon^\star$ algorithm~\cite{song2018} to consider the energy constraints on autonomous vehicles. Therefore, we first present a review of the $\epsilon^\star$ algorithm and then discuss the new additions in the $\epsilon^\star$+ algorithm. 

\begin{figure*}[t]
    \centering
    \includegraphics[width=0.95\textwidth]{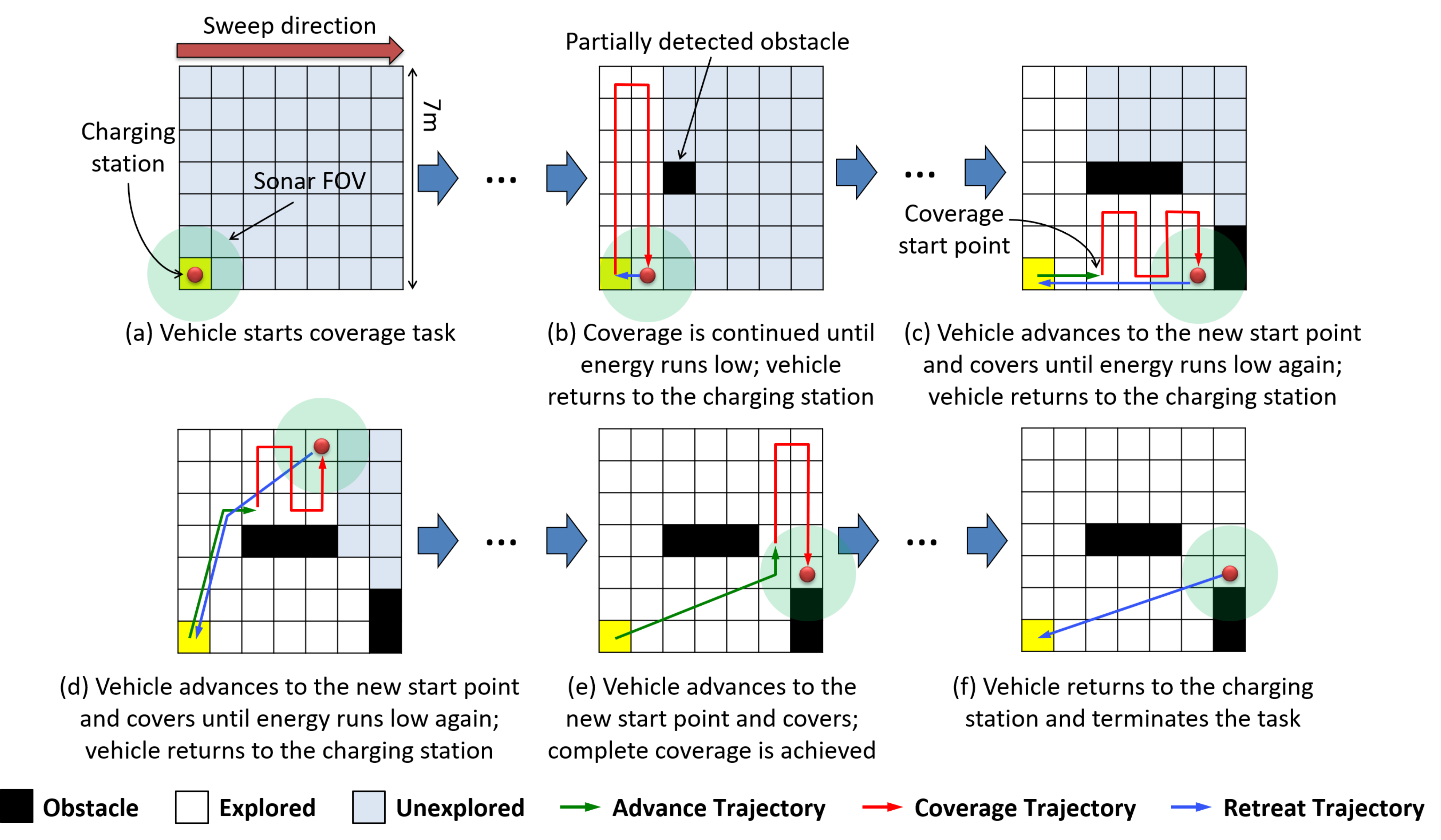}
    \caption{An example illustrating the execution of the $\epsilon^\star$+ algorithm in an environment containing unknown obstacles.}
   \label{fig:example} %\vspace{-10pt}
\end{figure*}

\subsection{Review of the $\epsilon^\star$ Algorithm}
This paper utilizes an online coverage path planning algorithm called $\epsilon^\star$~\cite{song2018}. This algorithm works in unknown  environments, where the coverage trajectory must be adjusted \textit{in situ}~\cite{GRP09} when new obstacle information is available. The key benefits of the $\epsilon^\star$ algorithm include low computation burden for real-time application, complete coverage guarantee, back and forth search, and absence of local extremum. It also differs from the cellular decomposition based methods and does not require detection of critical points. During the coverage process, the $\epsilon^\star$ algorithm takes the sensor feedback from the vehicle and updates the environmental information as time-varying potentials on Multiscale Adaptive Potential Surfaces (MAPS), which act as guidance surfaces for decision making. The algorithm utilizes
an Exploratory Turing Machine (ETM) as a supervisor for making navigation decisions. MAPS  enable the vehicle to incrementally use the environmental information for path planning and  escaping from local extremum. To build MAPS, a hierarchical multiscale tiling (MST) is constructed on the search area $\mathcal{A}$ in total $K+1$ layers (i.e., Level $0,\dots,K\in\mathbb{N}$), representing the search area with various resolutions. Specifically, the tiling $\mathcal{T}$ indicates the finest level with the highest resolution and is referred as $\mathcal{T}^0$.

%The detailed procedure to construct such structure is presented in~\cite{song2018}.

\begin{figure*}[t]
    \centering
    \includegraphics[width=0.8\textwidth]{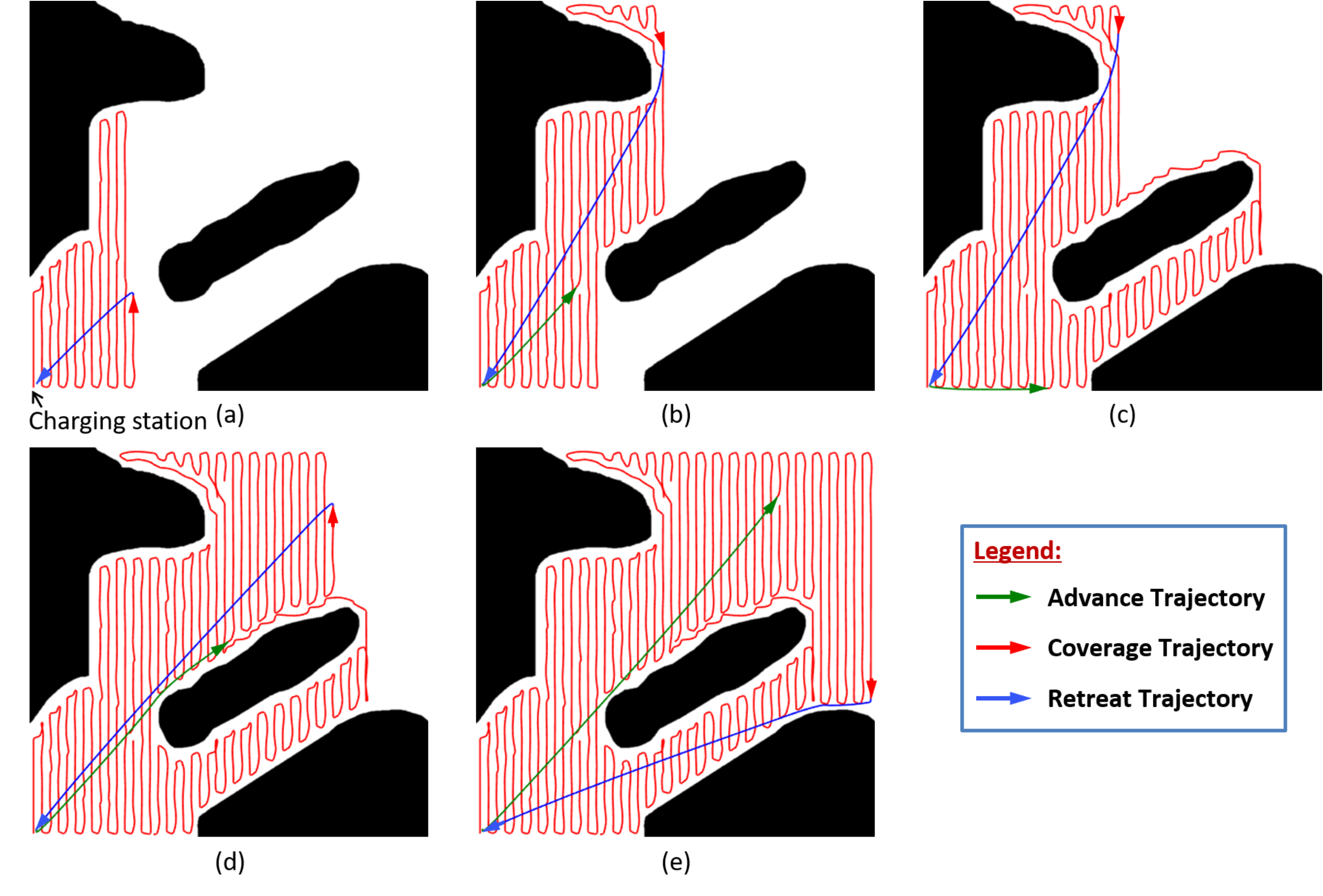}
    \caption{The vehicle trajectory showing complete coverage of the search area of scenario 1.}
  \label{fig:trajectory1} %\vspace{-10pt}
\end{figure*}

By default, the $\epsilon^\star$ algorithm utilizes the MAPS at level $0$ to generate the coverage trajectory online. When the vehicle falls in a local extremum, the algorithm swithches to higher levels to escape from a local extremum. For modeling of MAPS at level $0$, the collected environmental information is encoded on each cell $\tau_{{\alpha}^0}\in\mathcal{T}^0$ based on its physical state $S_{\tau_{{\alpha}^0}} \in S=\{O,E,U\}$, where $O\equiv obstacle, E \equiv explored,$ and $U \equiv unexplored$. The states of all the cells are initialized with state $U$ as the search area is assumed to be \textit{a priori} unknown in the beginning. As the vehicle explores the search area, the cells traversed by its coverage trajectory are updated as $E$, while those occupied by detected obstacles are marked as $O$. Further, based on the encoded symbols, a potential surface is built by assigning a discrete potential to each $\tau_{{\alpha}^0}$ as follows:

\begin{equation}
    \mathcal{E}_{\tau_{{\alpha}^0}}(k) =
    \begin{cases}
        -1, &  \textrm{if} \ S_{\tau_{{\alpha}^0}}(k)= O \\
        0, &  \textrm{if} \ S_{\tau_{{\alpha}^0}}(k)=E \\
        B_{{\alpha}^0}, &  \textrm{if} \ S_{\tau_{{\alpha}^0}}(k)=U 
    \end{cases}
\end{equation}
where $S_{\tau_{{\alpha}^0}}(k) \in S$ is the state of $\tau_{{\alpha}^0}$ at time instant $k$. The first condition assigns a potential of -1 to $\tau_{{\alpha}^0}$, if it contains an obstacle. The second condition assigns a potential of 0 to $\tau_{{\alpha}^0}$, if it has been explored by the vehicle. The last condition assigns a potential of $B_{{\alpha}^0}$ to $\tau_{{\alpha}^0}$, if it is yet unexplored, where $B=\{B_{{\alpha}^0} \in \{1,\dots,B_{max}\}, {\alpha}^0 = 1,\cdots,|\mathcal{T}^0|\}$ is designed offline to have plateaus of equipotential surfaces along each column of the tiling. It assigns equal potential to the cells lying in the same column, while the potentials in different columns increase gradually from right to left.

\begin{figure*}[t]
    \centering
    \includegraphics[width=0.8\textwidth]{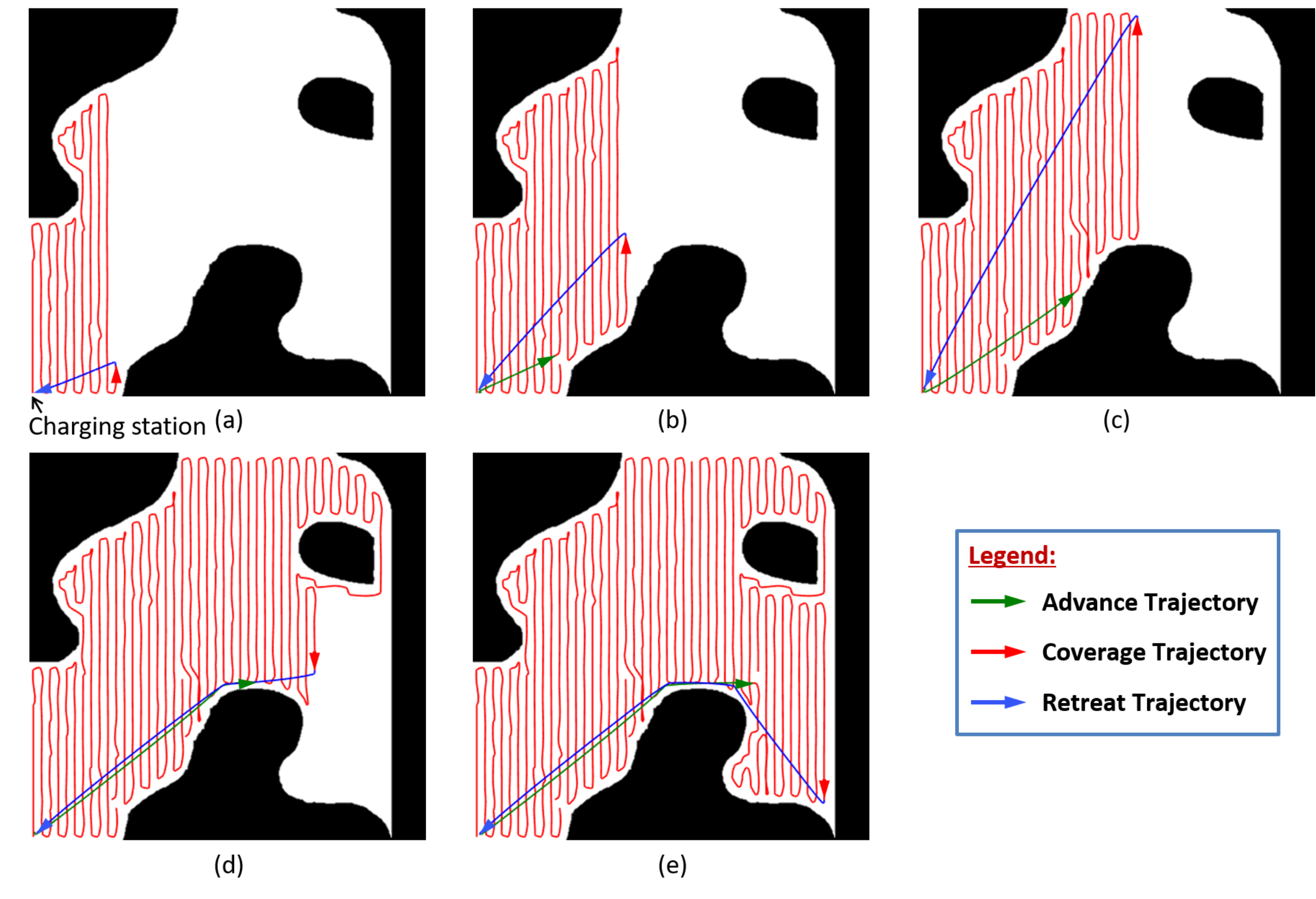}
    \caption{The vehicle trajectory showing complete coverage of the search area of scenario 2.}
  \label{fig:trajectory2} %\vspace{-10pt}
\end{figure*}

For navigation using the lowest level, the cell possessing the highest potential in the local neighborhood is set as the navigation goal. Note that there could be more than one cell with the highest potential, if they belong to an equipotential surface. In that case, the vehicle selects the cell $\tau_\mu$ which has the least cost $C_{\mu,\lambda}$ to reach it from the current cell $\tau_\lambda$. 

The potential surfaces at higher levels of MAPS are built  using a probabilistic approach. For escaping from local extrema, the ETM swithces to the higher levels of MAPS incrementally as needed and finds an unexplored cell as the waypoint; thus guaranteeing complete coverage.  

%In this paper, $C_{\mu,\lambda}$ is defined as the combination of travel cost and the turning cost along the straight path needed to reach the centroid of the cell $\tau_{\mu}$ from the current cell $\tau_\lambda$. 

\subsection{Novel Features of the $\epsilon^\star$+ Algorithm}
The $\epsilon^\star$+ algorithm considers energy-constraints of the autonomous vehicle during coverage. Thus, it enables the vehicle to retreat back to the charging station based on the remaining energy which is monitored throughout the coverage process. At all times, the remaining energy should be sufficient to allow the vehicle to retreat back to the charging station. Thus, to determine if the remaining energy is sufficient for retreat, the $\epsilon^\star$+ algorithm first computes the retreat trajectory. This is done as follows. Each time the navigation waypoint is computed by the $\epsilon^\star$ algorithm, the retreat trajectory is computed from this waypoint to the charging station, as the shortest path. Specifically, the visibility graph~\cite{lozano1979} is extracted from the latest symbolically encoded tiling structure. In this graph, the node set consists of the charging station, the computed navigation waypoint, and the vertices of the obstacles. Note that a vertex of the obstacle is the free cell lying on the obstacle boundary. The nodes of the graph are connected by edges excluding the edges which are obstructed by an obstacle or unknown cells. Then, the retreat path is obtained by performing the $\text{A}^{\star}$ search~\cite{hart1968} on the visibility graph.

Next, the expected energy consumption, for retreating back from the navigation waypoint to the charging station, is evaluated against the vehicle's remaining energy, such that it can maintain sufficient energy for returning back to the charging station after covering the navigation waypoint. 

\begin{defn}[Energy Consumption]\label{define:energy}
The energy consumption of the vehicle is modeled as proportional to the trajectory length~\cite{shnaps2016} for the advance and retreat segments, while it is twice this amount for the coverage segment.
\end{defn}

The vehicle computes the remaining energy at the current cell as the initial energy minus the energy consumed by the trajectory taken so far. If the remaining energy is less than the expected energy consumption for covering the next navigation waypoint and then retreating back to the charging station, then the vehicle returns back to the charging station. After recharging, the vehicle starts the coverage process fresh with an updated map which is partially covered and partialy known. At the charging station, the vehicle uses the MAPS structure of the $\epsilon^\star$ algorithm to find a nearby unexplored point. Specifically, the $\epsilon^\star$ algorithm switches to a coarse level on the MAPS until it finds an unexplored coarse cell with the highest positive potential. Subsequently, within this coarse cell, the closest unexplored cell at the finest level is selected as the new coverage start point. Then the advance trajectory from the charging station to this unexplored cell is computed using the visibility graph and the $\text{A}^{\star}$ search, as described above. 

Fig.~\ref{fig:example} shows an example of the $\epsilon^\star$+ algorithm for online energy-constrained CPP in an unknown environment. The vehicle with the total energy capacity of $14$ units, and sonar field of view (FOV) of radius $1.5 \text{m}$, starts at the charging station marked with yellow. In the coverage segment, the vehicle consumes $1$ unit of energy for $1 \text{m}$ of coverage per cell, while in the advance and retreat segments, it consumes 0.5 units of energy for $1 \text{m}$ of travel. In the beginning, the environment is initialized as unexplored as it is \textit{a priori} unknown, as shown in Fig.~\ref{fig:example} (a). Then, the vehicle covers the search area following the back-and-forth motion while discovering the obstacles continuously. When the energy runs low, the vehicle returns to the charging station along the retreat trajectory, as shown in Fig.~\ref{fig:example} (b).
After recharging, the vehicle moves to the closest coverage start point along the advance trajectory, as shown in Fig.~\ref{fig:example} (c). Then, the vehicle continues the coverage until energy runs low again and retreats back. This process repeats until the complete coverage is achieved, as shown in Fig.~\ref{fig:example} (e). Finally, the vehicle returns to the charging station and terminates the task, as shown in Fig.~\ref{fig:example}(f).

Table.~\ref{tab:feature} presents the qualitative comparison between the key features of the $\epsilon^\star$+ and BPC algorithms~\cite{shnaps2016}.

 %$\epsilon^{\star}$ algorithm utilizes an Exploratory Turing Machine (ETM) as a supervisor to guide the vehicle with adaptive navigation decisions. However, it does not consider the limited energy capacity of the vehicle, thus leading to incomplete coverage.  

\begin{table}[!t]{}
%\captionsetup{justification=centering}
%\captionsetup{labelsep=none}
%\captionsetup{labelsep=newline}
%\caption {\sc {Comparison of Key Features of CT-CPP with TF-CPP}} \label{tab:feature}
\caption { Comparison of key features of the $\epsilon^*$+ and BPC Algorithms}\label{tab:feature}
\centering
 \begin{tabular}{l l l} 
 \toprule
  &  $\epsilon^\star$+ algorithm &  BPC algorithm~\cite{shnaps2016} \\ 
 \hline \vspace{-4pt}
\tabincell{l}{Environment} &\tabincell{l} Planar surface
 &\tabincell{l}Planar surface\\ 
 \specialrule{0em}{3pt}{3pt}\vspace{-4pt}
  Path Pattern &\tabincell{l}{Back and forth path} & \tabincell{l}{Circular contour} \\
 \specialrule{0em}{3pt}{3pt} %\vspace{-4pt}
  Approach &\tabincell{l}{Uses ETM as \\ a supervisor with \\ online energy monitor}  & \tabincell{l}{ Relies on the construction \\ of equipotential contours,\\ split cells and corridors}\\
    \specialrule{0em}{3pt}{3pt} %\vspace{-4pt}
  Remarks &\tabincell{l}{Produces easy-to-follow \\ back-and-forth path \\ Selects the closest cell\\ as the coverage start \\ point after each retreat \\ to avoid longer travel to \\ the previous waypoint}  & \tabincell{l}{Generates difficult-to-follow \\ contour path. Continues the \\ coverage of current corridor \\ after recharging if it is not \\ completely covered which \\ can produce longer travel to \\ the previous waypoint}\\
 
 %\hline
 \toprule
 \end{tabular}
 \vspace{-10pt}
 \end{table}

\section{Results}\label{sec:results}

The performance of the proposed algorithm is validated on a high-fidelity simulator called Player/Stage~\cite{gerkey2003}, which is a software tool for visualization and simulation of the robotic tasks. Player provides a software base whose libraries contain models of different types of robots and sensors while Stage is a highly configurable robot simulator. In this work, the vehicle is considered to have the total energy capacity of $320$ units. It is equipped with a sonar sensor with a maximum range of $5\text{m}$ for obstacle detection. Two complex scenarios of $50\text{m}\times50\text{m}$ search areas containing multiple obstacles are used for testing and validation. Each of these scenarios is partitioned into a $50\times50$ tiling structure, where each cell is of size $1\text{m}\times1\text{m}$.

Figs.~\ref{fig:trajectory1} and~\ref{fig:trajectory2} show the results of the proposed $\epsilon^\star$+ algorithm for two different scenarios, respectively. In both scenarios, the battery charging station is located at the bottom left corner. The two figures show the vehicle's coverage trajectory at different iterations. In the beginning, the vehicle starts the coverage of the search area at the charging station, as shown in Figs.~\ref{fig:trajectory1}(a) and~\ref{fig:trajectory2}(a). As the time-invariant field energy $B$ is predefined with a higher value at the left, the vehicle covers from the left to the right gradually. Since the cells lying in the same column have equal potential, the vehicle tends to cover in a column-wise manner. During the coverage process, the vehicle detects the obstacles, updates the remaining energy, and computes the expected energy consumption from the navigation waypoint to the charging station. When the energy runs low, the vehicle returns to the charging station following the computed retreat trajectory marked with blue, as shown in Figs.~\ref{fig:trajectory1}(a) and \ref{fig:trajectory2}(a). After fully recharging, the vehicle returns to a nearby coverage start point computed by the $\epsilon^\star$ algorithm along the advance trajectory marked with green, as shown in Figs.~\ref{fig:trajectory1}(b) and \ref{fig:trajectory2}(b). Then, the vehicle continues the coverage process until the energy runs low again and at that point the vehicle retreats back again. This process is repeated until the complete coverage is achieved, as shown in Figs.~\ref{fig:trajectory1}(e) and ~\ref{fig:trajectory2}(e). Finally, Figs.~\ref{fig:trajectory1}(e) and~\ref{fig:trajectory2}(e) present the complete coverage of the whole area, thus proving the effectiveness of the proposed algorithm.

\section{Conclusions and Future Work}
\label{sec:conclusions}
In this paper, a novel algorithm called $\epsilon^\star$+ is proposed for efficient online coverage of unknown environments using energy-constrained vehicles. The $\epsilon^\star$+ algorithm is an extension of the $\epsilon^\star$ algorithm to account for the energy constraints of the vehicle. The $\epsilon^\star$+ algorithm monitors the remaining energy of the vehicle throughout the coverage process and plans the retreat trajectory to the charging station based on the remaining energy and then after recharging it plans the advance trajectory  to a new waypoint to restart the coverage. Thus, the vehicle is able to autonomously navigate in an unknown environment while avoiding obstacles and maintaining the sufficient energy for returning back to the charging station as needed. Its performance is validated on two scenarios in Player/Stage. 

Future work includes extending the energy-constrained CPP problem to consider kinematic constraints~\cite{shen2019}, multi-agent systems~\cite{song2020} and  risk~\cite{song2018t}. Variable-speed and acceleration constraints may be considered to make the vehicle motion more realistic. Finally, a sample-based approach for CPP will be investigated to enable more faster and adaptive waypoint selection for real-time decision in dynamic environments.

\bibliographystyle{IEEEtran}
\bibliography{reference}

% Generated by IEEEtran.bst, version: 1.14 (2015/08/26)
\begin{thebibliography}{10}
\providecommand{\url}[1]{#1}
\csname url@samestyle\endcsname
\providecommand{\newblock}{\relax}
\providecommand{\bibinfo}[2]{#2}
\providecommand{\BIBentrySTDinterwordspacing}{\spaceskip=0pt\relax}
\providecommand{\BIBentryALTinterwordstretchfactor}{4}
\providecommand{\BIBentryALTinterwordspacing}{\spaceskip=\fontdimen2\font plus
\BIBentryALTinterwordstretchfactor\fontdimen3\font minus
  \fontdimen4\font\relax}
\providecommand{\BIBforeignlanguage}[2]{{%
\expandafter\ifx\csname l@#1\endcsname\relax
\typeout{** WARNING: IEEEtran.bst: No hyphenation pattern has been}%
\typeout{** loaded for the language `#1'. Using the pattern for}%
\typeout{** the default language instead.}%
\else
\language=\csname l@#1\endcsname
\fi
#2}}
\providecommand{\BIBdecl}{\relax}
\BIBdecl

\bibitem{shen2016}
Z.~Shen, J.~Song, K.~Mittal, and S.~Gupta, ``An autonomous integrated system
  for 3-d underwater terrain map reconstruction,'' in \emph{OCEANS 2016
  MTS/IEEE Monterey}.\hskip 1em plus 0.5em minus 0.4em\relax IEEE, 2016, pp.
  1--6.

\bibitem{shen2017}
------, ``Autonomous 3-d mapping and safe-path planning for underwater terrain
  reconstruction using multi-level coverage trees,'' in \emph{OCEANS
  2017-Anchorage}.\hskip 1em plus 0.5em minus 0.4em\relax IEEE, 2017, pp. 1--6.

\bibitem{shen2017mastersthesis}
\BIBentryALTinterwordspacing
Z.~Shen, ``3-d coverage path planning for underwater terrain mapping,''
  Master's thesis, University of Connecticut, 2017. [Online]. Available:
  \url{https://opencommons.uconn.edu/gs_theses/1133}
\BIBentrySTDinterwordspacing

\bibitem{englot2013}
B.~Englot and F.~S. Hover, ``Three-dimensional coverage planning for an
  underwater inspection robot,'' \emph{The International Journal of Robotics
  Research}, vol.~32, no. 9-10, pp. 1048--1073, 2013.

\bibitem{mukherjee2011}
K.~Mukherjee, S.~Gupta, A.~Ray, and S.~Phoha, ``Symbolic analysis of sonar data
  for underwater target detection,'' \emph{IEEE Journal of Oceanic
  Engineering}, vol.~36, no.~2, pp. 219--230, 2011.

\bibitem{song2013}
J.~Song, S.~Gupta, J.~Hare, and S.~Zhou, ``Adaptive cleaning of oil spills by
  autonomous vehicles under partial information,'' in \emph{2013 OCEANS-San
  Diego}.\hskip 1em plus 0.5em minus 0.4em\relax IEEE, 2013, pp. 1--5.

\bibitem{acar2002}
E.~U. Acar and H.~Choset, ``Sensor-based coverage of unknown environments:
  Incremental construction of morse decompositions,'' \emph{The International
  Journal of Robotics Research}, vol.~21, no.~4, pp. 345--366, 2002.

\bibitem{gabriely2003}
Y.~Gabriely and E.~Rimon, ``Competitive on-line coverage of grid environments
  by a mobile robot,'' \emph{Computational Geometry}, vol.~24, no.~3, pp.
  197--224, 2003.

\bibitem{gonzalez2005}
E.~Gonzalez, O.~Alvarez, Y.~Diaz, C.~Parra, and C.~Bustacara, ``Bsa: a complete
  coverage algorithm,'' in \emph{Proceedings of the 2005 IEEE International
  Conference on Robotics and Automation}.\hskip 1em plus 0.5em minus
  0.4em\relax IEEE, 2005, pp. 2040--2044.

\bibitem{galceran2013}
E.~Galceran and M.~Carreras, ``A survey on coverage path planning for
  robotics,'' \emph{Robotics and Autonomous Systems}, vol.~61, no.~12, pp.
  1258--1276, 2013.

\bibitem{shen2019}
Z.~Shen, J.~P. Wilson, and S.~Gupta, ``An online coverage path planning
  algorithm for curvature-constrained {AUVs},'' in \emph{OCEANS 2019 MTS/IEEE
  SEATTLE}.\hskip 1em plus 0.5em minus 0.4em\relax IEEE, 2019, pp. 1--5.

\bibitem{song2018}
J.~Song and S.~Gupta, ``$\epsilon^\star$: An online coverage path planning
  algorithm,'' \emph{IEEE Transactions on Robotics}, vol.~34, no.~2, pp.
  526--533, 2018.

\bibitem{shnaps2016}
I.~Shnaps and E.~Rimon, ``Online coverage of planar environments by a battery
  powered autonomous mobile robot,'' \emph{IEEE Transactions on Automation
  Science and Engineering}, vol.~13, no.~2, pp. 425--436, 2016.

\bibitem{wei2018}
M.~Wei and V.~Isler, ``Coverage path planning under the energy constraint,'' in
  \emph{2018 IEEE International Conference on Robotics and Automation
  (ICRA)}.\hskip 1em plus 0.5em minus 0.4em\relax IEEE, 2018, pp. 368--373.

\bibitem{GRP09}
S.~Gupta, A.~Ray, and S.~Phoha, ``Generalized ising model for dynamic
  adaptation in autonomous systems,'' \emph{Euro. Phys. Lett.}, vol.~87, no.~1,
  p. 10009, 2009.

\bibitem{lozano1979}
T.~Lozano-P{\'e}rez and M.~A. Wesley, ``An algorithm for planning
  collision-free paths among polyhedral obstacles,'' \emph{Communications of
  the ACM}, vol.~22, no.~10, pp. 560--570, 1979.

\bibitem{hart1968}
P.~E. Hart, N.~J. Nilsson, and B.~Raphael, ``A formal basis for the heuristic
  determination of minimum cost paths,'' \emph{IEEE transactions on Systems
  Science and Cybernetics}, vol.~4, no.~2, pp. 100--107, 1968.

\bibitem{gerkey2003}
B.~Gerkey, R.~T. Vaughan, and A.~Howard, ``The player/stage project: Tools for
  multi-robot and distributed sensor systems,'' in \emph{Proceedings of the
  11th international conference on advanced robotics}, vol.~1, 2003, pp.
  317--323.

\bibitem{song2020}
J.~Song and S.~Gupta, ``Care: Cooperative autonomy for resilience and
  efficiency of robot teams for complete coverage of unknown environments under
  robot failures,'' \emph{Autonomous Robots}, vol.~44, no.~3, pp. 647--671,
  2020.

\bibitem{song2018t}
J.~Song, S.~Gupta, and T.~A. Wettergren, ``T$^\star$: Time-optimal risk-aware
  motion planning for curvature-constrained vehicles,'' \emph{IEEE Robotics and
  Automation Letters}, vol.~4, no.~1, pp. 33--40, 2018.

\end{thebibliography}
\end{document}